\documentclass{article}

\usepackage{microtype}
\usepackage{graphicx}
\usepackage{subcaption}
\usepackage{booktabs} % for professional tables

\usepackage{hyperref}

\usepackage[preprint]{icml2026}

\usepackage{amsmath}
\usepackage{amssymb}
\usepackage{mathtools}
\usepackage{amsthm}
\usepackage{multirow} % Added for vertical centering
\usepackage{arydshln} % Added for dashed lines (must be loaded after booktabs)
\usepackage{xcolor}
\usepackage{array}

\usepackage[capitalize,noabbrev]{cleveref}

\theoremstyle{plain}

\theoremstyle{definition}

\theoremstyle{remark}

\usepackage[textsize=tiny]{todonotes}

\icmltitlerunning{MoVE: Mixture of Value Embeddings}

\begin{document}

\twocolumn[
  \icmltitle{MoVE: Mixture of Value Embeddings \\
    A New Axis for Scaling Parametric Memory in Auto-Regressive Models}



  \icmlsetsymbol{equal}{*}

  \begin{icmlauthorlist}
    \icmlauthor{Yangyan Li}{comp}
    %\icmlauthor{Firstname1 Lastname1}{equal,yyy}
    %\icmlauthor{Firstname2 Lastname2}{equal,yyy,comp}
    %\icmlauthor{Firstname3 Lastname3}{comp}
    %\icmlauthor{Firstname4 Lastname4}{sch}
    %\icmlauthor{Firstname5 Lastname5}{yyy}
    %\icmlauthor{Firstname6 Lastname6}{sch,yyy,comp}
    %\icmlauthor{Firstname7 Lastname7}{comp}
    %\icmlauthor{Firstname8 Lastname8}{sch}
    %\icmlauthor{Firstname8 Lastname8}{yyy,comp}
  \end{icmlauthorlist}

  \icmlaffiliation{comp}{Ant Group, Hangzhou, China}
  %\icmlaffiliation{yyy}{Department of XXX, University of YYY, Location, Country}
  %\icmlaffiliation{comp}{Company Name, Location, Country}
  %\icmlaffiliation{sch}{School of ZZZ, Institute of WWW, Location, Country}

  %\icmlcorrespondingauthor{Firstname1 Lastname1}{first1.last1@xxx.edu}
  %\icmlcorrespondingauthor{Firstname2 Lastname2}{first2.last2@www.uk}
  \icmlcorrespondingauthor{Yangyan Li}{yangyan.lyy@antgroup.com}

  \icmlkeywords{Auto-Regressive Modelling, Transformer, Attention, Model Scaling}

  \vskip 0.3in
]



\printAffiliationsAndNotice{}  % no special notice (required even if empty)

\begin{abstract}
\looseness=-1
Autoregressive sequence modeling stands as the cornerstone of modern Generative AI, powering results across diverse modalities ranging from text generation to image generation. However, a fundamental limitation of this paradigm is the rigid structural coupling of model capacity to computational cost: expanding a model's parametric memory---its repository of factual knowledge or visual patterns---traditionally requires deepening or widening the network, which incurs a proportional rise in active FLOPs.
In this work, we introduce \textbf{MoVE (Mixture of Value Embeddings)}, a mechanism that breaks this coupling and establishes a new axis for scaling capacity. MoVE decouples memory from compute by introducing a global bank of learnable value embeddings shared across all attention layers. For every step in the sequence, the model employs a differentiable soft gating mechanism to dynamically mix retrieved concepts from this bank into the standard value projection.
This architecture allows parametric memory to be scaled independently of network depth by simply increasing the number of embedding slots. We validate MoVE through strictly controlled experiments on two representative applications of autoregressive modeling: Text Generation and Image Generation. In both domains, MoVE yields consistent performance improvements over standard and layer-wise memory baselines, enabling the construction of ``memory-dense'' models that achieve lower perplexity and higher fidelity than their dense counterparts at comparable compute budgets.
\end{abstract}
\section{Introduction}

The Transformer architecture \citep{vaswani2017attention} and the autoregressive modeling paradigm have expanded beyond their origins in Natural Language Processing (NLP) to become the unified engine of Generative AI. Whether predicting the next subword in a sentence or the next discrete visual code in an image \citep{llamagen}, the underlying principle remains the same: modeling the joint probability distribution of a sequence via conditional ``next-token prediction.''

Across these modalities, performance is governed by scaling laws \citep{kaplan2020scaling, hoffmann2022training}, which traditionally impose a rigid tradeoff: increasing a model's capacity to store and utilize information (its parametric memory) requires increasing the number of dense parameters by deepening or widening the network. This creates a structural bottleneck where scaling memory capacity incurs a linear penalty in training and inference FLOPs. Breaking this coupling is a significant area of interest for advancing GenAI, as it would allow for ``memory-dense'' models that possess vast encyclopedic or visual knowledge without the prohibitive compute cost of massive dense networks.

\begin{figure}[t]
\begin{center}
\centerline{\includegraphics[width=\columnwidth]{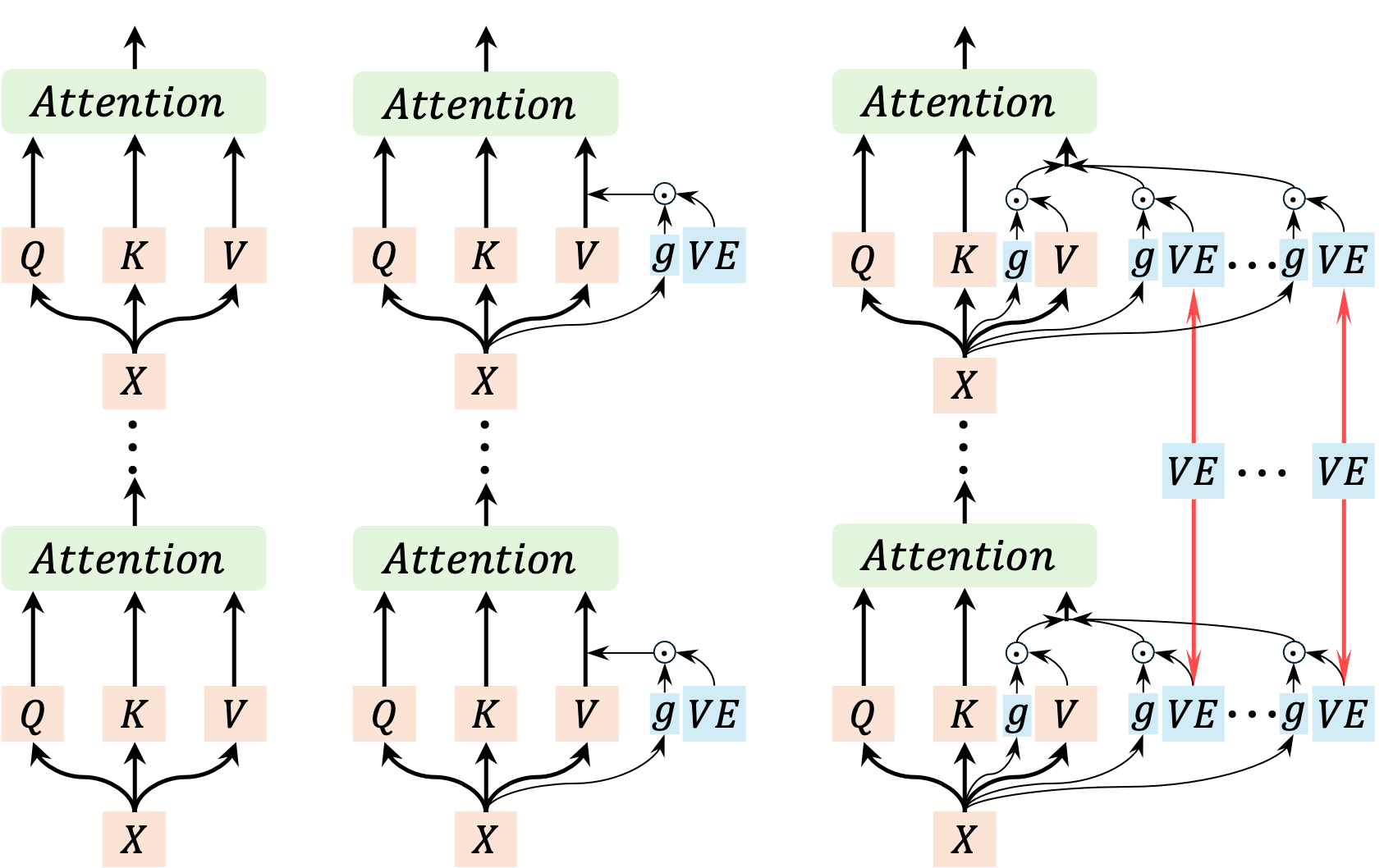}}
\caption{Architectural Comparison. \textbf{Left:} Standard Attention, where the Value stream is a linear projection of the local hidden state $X$. \textbf{Center:} Layer-wise Value Embeddings (LaVE), a baseline that augments attention with a layer-specific memory vector gated by the input. \textbf{Right:} \textbf{MoVE (Ours)}, which decouples memory from depth by introducing a global bank of value embeddings. Crucially, both layers fetch value embeddings from this central shared bank, employing a differentiable router to dynamically mix these global concepts into the value stream.}
\label{fig:move_arch}
\end{center}
\end{figure}

In this work, we introduce \textbf{MoVE (Mixture of Value Embeddings)}, a modality-agnostic mechanism designed to address this efficiency challenge. MoVE adapts the spirit of memory augmentation \citep{sukhbaatar2019augmenting} into a fully differentiable, dense form compatible with standard Transformers (see Figure \ref{fig:move_arch}). MoVE operates by augmenting the Attention mechanism's \textbf{Value stream}---which mechanistic interpretability research identifies as a key carrier of semantic content \citep{geva2021transformer}---with a global bank of learnable ``concept vectors'' shared across all attention layers. For each step in the sequence, the model computes a soft gating coefficient for every slot in this bank, dynamically mixing these retrieved concepts with the standard value projection. This allows the model to potentially offload the storage of static information (e.g., definitions of words, textures of objects) to the embedding bank, freeing up the dense attention layers for complex reasoning and structure building.

We validate the efficacy of MoVE through strictly controlled comparisons on the two most prominent pillars of autoregressive modeling:

\textbf{Text Generation.} We train Large Language Models (LLMs) \citep{brown2020language} using standard open-source datasets. We compare MoVE against standard and strong baselines, maintaining identical hyperparameters to isolate the architectural differences.

\textbf{Image Generation.} We integrate MoVE into the \textbf{LlamaGen} framework \citep{llamagen}, a representative autoregressive image generator. We show that MoVE improves the fidelity of generated images compared to the vanilla LlamaGen baseline, demonstrating that the mechanism's benefits transfer effectively to visual memory.

Furthermore, we demonstrate the architectural versatility of MoVE by extending it to \textbf{Multi-Head Latent Attention (MLA)} \citep{deepseekv2}. MLA is an optimized architecture that compresses Key-Value heads into a low-rank latent vector for inference efficiency. We show that MoVE extends seamlessly to this setting by injecting memory directly into the compressed latent space, enhancing capacity without disrupting the efficiency gains of MLA.

By decoupling parametric memory from computational depth, MoVE establishes a distinct new axis for scaling. It serves as a general-purpose booster for autoregressive modeling, allowing practitioners to scale model capacity by simply increasing the number of mixture components. This enables the construction of ``memory-dense'' generative models that are smarter and richer in knowledge without the prohibitive cost of deepening the reasoning engine.
\section{Related Work}

\looseness=-1
\textbf{Unified Autoregressive Modeling.}
The autoregressive ``next-token prediction'' paradigm has proven to be a universal framework for generative AI. In NLP, this powers the dominant GPT series \citep{brown2020language}, where performance scales predictably with compute.
This paradigm has successfully transferred to Computer Vision, where works like ImageGPT \citep{chen2020generative}, VQGAN \citep{esser2021taming}, and Parti \citep{yu2022scaling} demonstrate that visual representations can be learned via sequences of discrete tokens. Most recently, frameworks like \textbf{LlamaGen} \citep{llamagen} show that standard Transformers—without modality-specific biases—achieve effective synthesis purely through autoregressive scaling. This convergence implies that architectural innovations such as MoVE are inherently modality-agnostic.

\looseness=-1
\textbf{Scaling Laws and Efficient Architectures.}
Model performance is governed by rigorous scaling laws \citep{kaplan2020scaling, hoffmann2022training}. However, traditional dense scaling requires a linear increase in training and inference FLOPs. To circumvent this, research has pivoted toward decoupled architectures. Universal Transformers \citep{dehghani2018universal} and ALBERT \citep{lan2019albert} attempted to decouple size from compute via cross-layer parameter sharing. While parameter-efficient, they do not inherently increase the model's \emph{memory capacity} relative to compute. MoVE explores an alternative direction by introducing a dedicated, scalable embedding bank that adds capacity without adding depth.
Specifically regarding attention efficiency, \textbf{Multi-Head Latent Attention (MLA)}, introduced in DeepSeek-V2 \citep{deepseekv2}, addresses the memory bottleneck of the KV cache by projecting Keys and Values into a low-rank latent vector. Our work is complementary to MLA; while MLA reduces the memory footprint of inference, MoVE increases the parametric memory capacity of the model. We show that these two goals can coexist by applying MoVE directly to the compressed latent space.

\textbf{Conditional Computation and Mixture-of-Experts.}
A prominent success in decoupling parameters from compute is the Mixture-of-Experts (MoE) paradigm \citep{shazeer2017outrageously, lepikhin2020gshard, jiang2024mixtral}, which routes tokens to sparse subsets of FFN ``experts.'' While successful, standard MoE relies on \textit{hard routing} (top-$k$), which often introduces training instability.
Recent work on \textit{Soft MoE} \citep{puigcerver2023soft} solves this by performing a differentiable weighted average of experts. MoVE adopts a similar philosophy but applies it to the Parameter space rather than the Token space. Unlike Soft MoE, which mixes \emph{tokens} into expert slots, MoVE mixes \emph{expert vectors} (value embeddings) into the token's representation. This allows MoVE to retain the standard autoregressive formulation while distinctively injecting the benefits of conditional computation.

\textbf{Memory-Augmented Neural Networks (MANNs).}
A parallel lineage augments networks with explicit memory. Memory Networks \citep{weston2014memory, sukhbaatar2015end} introduced reading from external banks. \citet{lample2019large} scaled this with Product Key Memory (PKM) using sparse retrieval.
The closest precursor to our work is \textit{Persistent Memory} \citep{sukhbaatar2019augmenting}, which concatenated static learnable vectors to the Key and Value matrices. However, Persistent Memory vectors are \textit{static}---they are attended to via the standard mechanism, forcing them to compete with context tokens for attention mass. MoVE builds upon this by making the memory access \textit{dynamic} and \textit{input-dependent} via a dedicated routing head ($x W_g$), decoupling the ``retrieval'' of global concepts from the ``attention'' to local context.
More recently, \citet{behrouz2025titans} introduced \textit{Titans}, which learns a neural memory module at test-time. MoVE differs by focusing on scaling \textit{parametric} (training-time) memory, offering a distinct way to store more encyclopedic knowledge in the weights themselves.

\textbf{Value Stream Optimization.}
Recent mechanistic interpretability work highlights the Value stream as the carrier of semantic content \citep{geva2021transformer, meng2022locating}.
Building on this, \citet{zhou2024value} proposed ResFormer and SVFormer, demonstrating that Value embeddings can be shared across layers to reduce redundancy without hurting performance. MoVE leverages this shared parameter hypothesis. However, while SVFormer shares a single static projection to reduce parameters, MoVE shares a \textit{large, diverse bank} of embeddings to \emph{increase} capacity. MoVE can be viewed as a high-capacity, dynamic generalization of the SVFormer principle.
\section{Methodology}

\looseness=-1
We introduce \textbf{MoVE (Mixture of Value Embeddings)} to decouple parametric memory scaling from Transformer compute. By augmenting the attention Value stream with a shared, differentiable embedding bank, MoVE enables capacity scaling orthogonal to network depth and width.

\subsection{Preliminaries: Autoregressive Modeling}
\label{sec:preliminaries}

\looseness=-1
We consider the standard decoder-only Transformer architecture \citep{vaswani2017attention}, which has become the de facto backbone for generative modeling.
In this paradigm, a sequence of discrete tokens $w = (w_1, \dots, w_T)$ where $w_t \in \{1, \dots, N_{vocab}\}$ is first mapped to a sequence of continuous hidden states $X \in \mathbb{R}^{T \times d}$. This is achieved via a learnable \textbf{Input Embedding} matrix $W_I \in \mathbb{R}^{N_{vocab} \times d}$ combined with positional encodings to retain sequence order information.

The core workhorse of this architecture is the Multi-Head Attention (MHA) module, which enables the model to capture complex dependencies across the sequence. Within each block, the input state $X$ is projected into distinct subspaces for each head $h \in \{1, \dots, H\}$:
\begin{equation}
    Q^{(h)} = X W_Q^{(h)}, \quad K^{(h)} = X W_K^{(h)}, \quad V^{(h)} = X W_V^{(h)}
\end{equation}
The self-attention mechanism then aggregates information globally by computing a weighted sum of values, where the weights are determined by the compatibility between queries and keys: $Y^{(h)} = \text{Softmax}(\frac{Q^{(h)} (K^{(h)})^T}{\sqrt{d_h}}) V^{(h)}$. The outputs from all heads are concatenated and mixed via a final linear projection $W_O$. Since this mechanism drives the reasoning capabilities of the model, architectural improvements to the attention block can directly benefit the entire family of tasks that rely on this modeling paradigm.

The model is trained using the \textbf{autoregressive objective}, which treats generation as a sequential decision process. The goal is to maximize the likelihood of the next token $w_t$ given the preceding context $w_{<t}$. The network parameters $\theta$ are optimized by minimizing the negative log-likelihood over the dataset: $\mathcal{L}(\theta) = - \sum_{t=1}^{T} \log P(w_t \mid w_{<t}; \theta)$.

\subsection{MoVE: Mixture of Value Embeddings}

MoVE intervenes in the generation of the Value tensors $V^{(h)}$. Standard value projections must encode all potential associations of a token into a single vector superposition. MoVE provides a bank of distinct embeddings and allows the model to dynamically select the relevant primitives for each attention head. We visualize the specific architectural differences between Standard Attention, LaVE (a strong baseline), and our proposed MoVE mechanism in Figure \ref{fig:move_arch}.

\textbf{Value Embedding Bank.}
We define a learnable global tensor $\mathcal{E} \in \mathbb{R}^{N_{vocab} \times M \times d}$, where $M$ is the number of embedding slots per token. Intuitively, this tensor represents a shared repository of primitives---whether they be linguistic definitions in text or visual attributes in image generation.
For a sequence of token indices $w$, we retrieve the memory values and reshape them to align with the attention heads:
\begin{equation}
    \mathcal{M}_t = \mathcal{E}[w_t] \in \mathbb{R}^{M \times H \times d_h}
\end{equation}
Here, $\mathcal{M}_{t, i}^{(h)}$ represents the $i$-th latent attribute of token $w_t$ specifically tailored for the subspace of head $h$.

\textbf{Distinction from Input Embeddings.}
Crucially, the Value Embedding bank $\mathcal{E}$ is distinct from the standard Input Embeddings $W_I$ defined in Sec.~\ref{sec:preliminaries}. In a standard Transformer, $W_I$ must serve a dual purpose: it acts as the source for generating attention patterns (via $Q$ and $K$) and as the carrier of semantic content (via $V$). MoVE decouples these roles. The Input Embeddings $W_I$ remain compact to facilitate efficient routing, while the Value Embeddings $\mathcal{E}$ provide a high-capacity, multi-slot repository of ``definitions'' strictly for the Value stream. This allows the model to store vast encyclopedic knowledge in $\mathcal{E}$ without expanding the dimension $d$ used for reasoning.

\textbf{Routing and Gating.}
Different attention heads often specialize in different functions. Therefore, we employ a \textbf{per-head router}. A projection $W_G \in \mathbb{R}^{d \times H(M + 1)}$ computes logits $Z \in \mathbb{R}^{T \times H \times (M + 1)}$ from the hidden state $X$.

We apply a scaled sigmoid activation to map the logits to the range $(0, 2)$:
\begin{equation}
    g_{t, i}^{(h)} = 2 \cdot \sigma(Z_{t, i}^{(h)})
\end{equation}
where $\sigma$ is the standard sigmoid function. The scalar factor of 2 initializes the gates (at $z=0$) to $1.0$, representing a neutral identity mixing, while allowing the model to amplify a signal (up to $2.0$) or suppress it (down to $0.0$).

\textbf{Value Mixing.}
The final value tensor for head $h$, $V_{\mathcal{S}}^{(h)}$, is a weighted sum of the standard projection and the retrieved global embeddings:
\begin{equation}
    V_{\mathcal{S}}^{(h)} = g_{t, 0}^{(h)} \odot V^{(h)} + \sum_{i=1}^{M} g_{t, i}^{(h)} \odot \mathcal{M}_{t, i}^{(h)}
\end{equation}

The first term (index 0), weighted by $g_{t, 0}^{(h)}$, corresponds to the standard dense projection $V^{(h)}$. Empirical results indicate that this term is critical for model performance. It serves as a base representation derived from the local context, which the retrieved memories then augment or refine. By learning to scale this term, the model can dynamically balance between contextual information (from $V^{(h)}$) and global parametric knowledge (from $\mathcal{M}$). Crucially, in the context of autoregressive inference, $V_{\mathcal{S}}^{(h)}$ serves as the definitive Value tensor; it is computed once at the current step and subsequently stored in the KV cache, ensuring that the overhead of memory retrieval is incurred only at generation time and not repeatedly during attention.

By sharing the embedding bank $\mathcal{E}$ across all attention layers, MoVE transforms the network into a collaborative system. Gradients from the shallowest feature detectors and the deepest reasoning heads flow back into the same shared tensor $\mathcal{E}$. This creates ``gradient highways'' that allow concepts to be learned rapidly and robustly, effectively densifying the signal for parameters that would otherwise be sparse and isolated in a layer-wise configuration.

\subsection{Analysis: The Cost of Intelligence vs. Memory}

MoVE offers an alternative scaling economy where memory is cheap and intelligence (compute) remains fixed.

\textbf{Parameter Efficiency via Indexing.}
MoVE introduces $N_{vocab} \cdot M \cdot d$ parameters. However, these are accessed via sparse indexing (loading only the relevant $M$ vectors for the active tokens), bypassing the memory bandwidth costs associated with dense FFN layers of equivalent size. Furthermore, because $\mathcal{E}$ is shared across all $L$ attention layers, the amortized parameter cost per layer decreases as the network deepens.

\textbf{Minimal Compute Overhead.} We quantify the computational cost by analyzing the floating-point operations (FLOPs) required per token. A standard Transformer block incurs a baseline cost of $C_{std} \approx 24d^2 + 4Td$, dominated by the dense attention and feed-forward projections. The overhead introduced by MoVE arises solely from the routing projection $W_G$, consuming $C_{move} \approx 2dH(M+1)$. The relative overhead ratio is:
\begin{equation}
    \frac{C_{move}}{C_{std}} \approx \frac{2 d H (M+1)}{24 d^2 + 4Td} = \frac{H(M+1)}{12d + 2T}
\end{equation}
For a large standard configuration (e.g., $d=2048, H=16$) with a memory bank of $M=32$ and context $T=2048$, this results in a negligible overhead of $\approx 1.8\%$. We refer the reader to \textbf{Appendix \ref{sec:appendix_flops}} for the detailed derivation of these computational bounds.
\section{Experiments}
\label{sec:experiments}

To validate the effectiveness of our proposed mechanism, we perform a comprehensive evaluation of MoVE across the two most prominent domains of autoregressive sequence modeling: Text Generation and Image Generation.

\looseness=-1
\textbf{Experimental Philosophy.}
Given resource constraints, our experiments operate at scales significantly smaller than commercial foundation models. Consequently, our goal is not to chase state-of-the-art benchmarks, but to demonstrate robust \emph{relative performance gains} through strictly controlled comparisons. We treat MoVE as the sole independent variable within the \texttt{nanochat} \citep{nanochat} and \texttt{LlamaGen} \citep{llamagen} frameworks. To isolate architectural benefits, we adhere to their standard configurations, ensuring that any necessary modifications (e.g., batch size adjustments for hardware utilization) are applied uniformly across all variants.

\subsection{Common Baselines and Scaling Settings}

To validate MoVE, we compare it against two consistent baseline architectures across modalities. Crucially, we define a unified scaling factor ``$\times$'' representing parametric memory capacity relative to network depth.

\textbf{1. Standard Transformer.} A dense, decoder-only Transformer. This represents the default architecture for both \texttt{nanochat} and \texttt{LlamaGen}.

\textbf{2. Layer-wise Value Embeddings (LaVE).} A strong baseline integrating layer-specific learnable memory. This architecture is adapted from the \texttt{modded-nanoGPT} repository \citep{moddednanogpt}.\footnote{The specific implementation of layer-wise value embeddings originates from the \texttt{modded-nanoGPT} community (contributed by @KoszarskyB) \citep{moddednanogpt}. Note that while the public reference implementation in \texttt{nanochat} computes the gate using only the first 32 channels of the input for efficiency, our LaVE baseline utilizes the full hidden dimension $d$ as the input to the gating projection $W_g^{(l)}$. This modification ensures strictly comparable expressivity between the LaVE and MoVE gating mechanisms in our controlled experiments.} In this setup, a distinct, independent embedding matrix $\mathcal{E}^{(l)} \in \mathbb{R}^{|\mathcal{V}| \times d_{kv}}$ is allocated to specific layers. \textbf{Crucially, these embeddings are layer-wise, non-shared, and strictly ``local'' to their associated layer.} For a token $x_t$ at layer $l$, the retrieved memory $\mathcal{M}_t^{(l)} = \mathcal{E}^{(l)}[x_t]$ is gated via a layer-specific projection $W_g^{(l)}$. To maintain consistency with the MoVE formulation, we denote the operation as:
\begin{equation}
    V_{\mathcal{S}}^{(l)} = V^{(l)} + g_{t}^{(l)} \odot \mathcal{M}_{t}^{(l)}
\end{equation}
where $V^{(l)}$ is the standard projection, $\mathcal{M}_{t}^{(l)}$ is the layer-specific retrieved memory, and the gate is computed as $g_{t}^{(l)} = 2 \cdot \sigma(x_t W_g^{(l)})$. Note that unlike MoVE (Eq. 6), LaVE does not gate the standard path $V^{(l)}$ and retrieves only a single vector per layer rather than a mixture. We utilize two settings for LaVE: \textbf{Default ($\times 1$)} integrates LaVE into the attention mechanism of half the Transformer blocks (indices $L-1, L-3, \dots$). \textbf{Dense ($\times 2$)} integrates LaVE into the attention mechanism of \emph{all} Transformer blocks. In both settings, memory capacity is strictly coupled to network depth ($M \propto L$).

\textbf{MoVE Scaling.}
For fair comparison, we align the configurations of MoVE and LaVE. \textbf{MoVE ($\times 1$)} uses a global bank of $M = L/2$ slots. \textbf{MoVE ($\times 2$)} uses a global bank of $M = L$ slots. However, unlike LaVE, MoVE decouples memory size from layer count. This allows us to scale further to \emph{$\times 4$} ($M = 2L$), \emph{$\times 8$} ($M = 4L$), and beyond without deepening the network. This capability to scale to ``super-dense'' memory regimes ($M > L$) is unique to MoVE.

\subsection{Application I: Text Generation}

\begin{table}[ht]
\caption{Text Generation Performance on FineWeb-Edu. \textbf{Val BPB} corresponds to the final dataset shard (\textbf{lower is better}). The \textbf{Scaling Gain} column reports the absolute reduction in bpb compared to the \emph{Standard} baseline of the same depth. MoVE demonstrates consistent gains in all of the three model scales.}
\label{tab:nlp_results}
\begin{center}
\begin{small}
\begin{sc}
\resizebox{\columnwidth}{!}{%
\begin{tabular}{lllrcc}
\toprule
Depth & Method & $M$ & Params & Val BPB $\downarrow$ & Gain $\uparrow$ \\
\midrule
\multirow{7}{*}{D12} & Standard & - & 186M & 0.838 & - \\[2pt] \cdashline{2-6} \addlinespace[2pt]
      & \multirow{2}{*}{+ LaVE} & $\times 1$ ($L/2$) & +302M & 0.822 & 0.016 \\
      &                         & $\times 2$ ($L$)   & +604M & 0.818 & 0.020 \\[2pt] \cdashline{2-6} \addlinespace[2pt]
      & \multirow{4}{*}{\textbf{+ MoVE}} & $\times 1$ ($L/2$) & +302M & 0.819 & 0.019 \\
      &                                  & $\times 2$ ($L$)   & +605M & 0.812 & 0.026 \\
      &                                  & $\times 4$ ($2L$)  & +1209M & 0.806 & 0.032 \\
      &                                  & $\times 8$ ($4L$)  & +2419M& \textbf{0.797} & \textbf{0.041} \\
\midrule
\multirow{6}{*}{D20} & Standard & - & 561M & 0.763 & - \\[2pt] \cdashline{2-6} \addlinespace[2pt]
      & \multirow{2}{*}{+ LaVE} & $\times 1$ ($L/2$) & +839M & 0.753 &0.010 \\
      &                         & $\times 2$ ($L$)   & +1678M  & 0.748 & 0.015 \\[2pt] \cdashline{2-6} \addlinespace[2pt]
      & \multirow{3}{*}{\textbf{+ MoVE}} & $\times 1$ ($L/2$) & +842M & 0.748 & 0.015 \\
      &                                  & $\times 2$ ($L$)   & +1683M  & 0.744 & 0.019 \\
      &                                  & $\times 4$ ($2L$)  & +3366M  & \textbf{0.739} & \textbf{0.024} \\
\midrule
\multirow{5}{*}{D32} & Standard & - & 1.88B & 0.693 & - \\[2pt] \cdashline{2-6} \addlinespace[2pt]
      & \multirow{2}{*}{+ LaVE} & $\times 1$ ($L/2$) & +2.15B & 0.683 & 0.010 \\
      &                         & $\times 2$ ($L$)   & +4.3B & 0.681 & 0.012 \\[2pt] \cdashline{2-6} \addlinespace[2pt]
      & \multirow{2}{*}{\textbf{+ MoVE}} & $\times 1$ ($L/2$) & +2.16B & 0.681 & 0.012 \\
      &                                  & $\times 2$ ($L$)   & +4.33B & \textbf{0.677} & \textbf{0.016} \\
\bottomrule
\end{tabular}
}
\end{sc}
\end{small}
\end{center}
\end{table}

We evaluate MoVE in the text domain using the \texttt{nanochat} framework \citep{nanochat}, adhering strictly to its default configurations to ensure a fair comparison.

\looseness=-1
\textbf{Models.} We perform experiments across three model depths defined by the framework. Note that \texttt{nanochat} scales model width proportionally to depth ($d_{model}=64 \times L$). Specifically, we train \textbf{D12} (12 layers, 768 dim, $\approx$186M params), \textbf{D20} (20 layers, 1280 dim, $\approx$561M params), and \textbf{D32} (32 layers, 2048 dim, $\approx$1.88B params).\footnote{Exact parameter counts for the standard baseline: D12: 185,597,976; D20: 560,988,200; D32: 1,879,048,256. While adding MoVE/LaVE slots increases the total parameter count, we maintain a fixed training duration for all variants to isolate architectural benefits.}

\textbf{Dataset and Tokenizer.} Models are pretrained on the \textbf{sample-100BT} subset of \emph{FineWeb-Edu} \citep{lozhkov2024fineweb-edu}, a curated high-quality dataset containing approximately 100 billion tokens processed into 1,823 shards, with the final shard reserved for evaluation. We utilize the framework's default BPE tokenizer with a vocabulary size of $V=65,536$, replicating the scheme used by \emph{GPT-4} \citep{openai2023gpt4} to balance efficient encoding with a compact embedding table.

\textbf{Training.} We train the D12 and D20 models for 30,000 steps, and the D32 model for 60,000 steps, with a global batch size of 524,288 tokens. This regime ensures that all model variants (Standard, LaVE, MoVE) are exposed to an identical volume of data, allowing for a rigorous comparison of their learning efficiency.\footnote{These step counts result in token-to-parameter ratios ranging from approx. 85:1 (D12) to 17:1 (D32). \citet{hoffmann2022training} establish a ratio of $\approx 20:1$ as the compute-optimal stopping point. Our setup trains the smaller models significantly beyond this frontier (into the regime of ``over-training''), while the larger models operate in its vicinity. However, our priority is not finding the absolute optimal duration for each scale, but aligning the training volume across variants. This strictly isolates the relative architectural gains of MoVE against the baselines under identical data exposure.}

\textbf{Metric.} Performance is evaluated on the reserved final shard of the dataset using \emph{Bits Per Byte (BPB)}. Unlike perplexity, BPB normalizes the loss by the raw UTF-8 byte count of the text. This provides a tokenizer-invariant measure of compression efficiency: $\text{BPB} = \mathcal{L}_{total} / (\ln(2) \cdot N_{bytes})$ where $\mathcal{L}_{total}$ is the total cross-entropy loss in nats and $N_{bytes}$ is the total number of UTF-8 bytes in the target sequence. Lower values indicate superior performance.

\textbf{Results.}
Table \ref{tab:nlp_results} summarizes the results. Both LaVE and MoVE provide tangible gains over the Standard Transformer. However, the results highlight two critical distinctions. First, \textbf{Parameter Efficiency}: MoVE uses parameters more effectively than LaVE. MoVE-$\times 1$ matches or exceeds the performance of LaVE-$\times 2$ in deeper models (e.g., 0.681 vs 0.681 at D32) and remains competitive at smaller scales (0.819 vs 0.818 at D12), despite having access to only half the number of memory slots. This suggests that the global shared memory mechanism allows for richer concept utilization than fragmented, layer-local storage. Second, \textbf{Scalability}: While LaVE is structurally bound by network depth and saturates quickly, MoVE demonstrates sustained scaling.

To better understand the internal dynamics of the MoVE mechanism, we conducted a qualitative visualization study on the gating behaviors of specific polysemous words. Our analysis reveals that the global memory slots exhibit distinct routing patterns corresponding to semantic, syntactic, and metaphorical shifts. We refer the reader to \textbf{Appendix \ref{sec:appendix_vis}} for a detailed discussion and visualization of these routing manifolds across different context lengths.

\subsection{Application II: Image Generation}

To validate the universality of MoVE, we integrate it into the \emph{LlamaGen} framework \citep{llamagen}. We treat image generation as analogous to language modeling, training autoregressive transformers on discrete visual tokens.

\textbf{Models.} We evaluate performance across two model scales from the framework: \textbf{GPT-B} (111M) and \textbf{GPT-L} (343M).\footnote{We also initiated experiments with the 1.4B parameter \textbf{GPT-XXL} model. While preliminary logs indicated encouraging reductions in training loss consistent with our other findings, we unfortunately overlooked that the checkpoint size of GPT-XXL exceeded the maximum file size limit of our configured checkpoint storage system. This resulted in a failure to save the final model weights, preventing us from running the generation benchmarks. These results will be included in a future revision.}

\begin{table}[ht]
\caption{Class-Conditional Image Generation on ImageNet (256x256). Following the protocol as described in LlamaGen, we prioritize \textbf{FID} as primary metric, while reporting \textbf{IS}, \textbf{Precision}, and \textbf{Recall} as secondary metrics for completeness. MoVE consistently outperforms the baselines on the primary FID metric.}
\label{tab:vision_results}
\begin{center}
\begin{small}
\begin{sc}
\resizebox{\columnwidth}{!}{%
\begin{tabular}{ll c ccc}
\toprule
\multirow{2}{*}{Size} & \multirow{2}{*}{Method} & \multicolumn{1}{c}{\textbf{Primary}} & \multicolumn{3}{c}{Secondary Metrics} \\
 \cmidrule(lr){3-3} \cmidrule(lr){4-6}
 & & \textbf{FID} $\downarrow$ & IS $\uparrow$ & Prec $\uparrow$ & Recall $\uparrow$ \\
\midrule
\multirow{3}{*}{GPT-B} & Standard & 6.53 & 167.3 & 0.83 & 0.43 \\
                       & + LaVE ($\times 1$)   & 6.11 & 175.9 & \textbf{0.84} & 0.43 \\
                       & \textbf{+ MoVE ($\times 1$)} & \textbf{5.62} & \textbf{191.7} & \textbf{0.84} & 0.43 \\
\midrule
\multirow{3}{*}{GPT-L} & Standard & 3.47 & \textbf{291.2} & 0.85 & 0.51 \\
                       & + LaVE ($\times 1$)   & 3.77 & 283.9 & \textbf{0.86} & 0.50 \\
                       & \textbf{+ MoVE ($\times 1$)} & \textbf{3.10} & 281.4 & 0.84 & \textbf{0.53} \\
\bottomrule
\end{tabular}
}
\end{sc}
\end{small}
\end{center}
\end{table}

\textbf{Dataset and Tokenizer.} Models are trained on \textbf{ImageNet-1K} (class-conditional). We employ the tokenizer pre-trained by the LlamaGen authors \citep{llamagen}.

\looseness=-1
\textbf{Training.} To fully leverage the 140GB HBM capacity of our training hardware, we increase the global batch size to 524,288 tokens, aligning it with our text generation experiments. All other hyperparameters---including the optimizer, learning rate schedule, and initialization---remain identical to the default LlamaGen configuration to ensure a fair comparison.

\looseness=-1
\textbf{Metrics.} Adhering to the standard evaluation protocol of \emph{LlamaGen} \citep{llamagen} and the broader community, we prioritize \textbf{Fréchet Inception Distance (FID)} \citep{heusel2017gans} as the primary benchmark for overall generative quality. To ensure strict comparability, we also report \textbf{Inception Score (IS)} \citep{salimans2016improved} and \textbf{Precision and Recall} \citep{kynkaanniemi2019improved} as secondary metrics.

\looseness=-1
\textbf{Results.}
Table \ref{tab:vision_results} summarizes the results.
%
%Consistent with the text domain, MoVE provides immediate improvements over the baselines.
%
For \textbf{GPT-B}, MoVE significantly reduces the main metric FID by \textbf{0.91} (from 6.53 to 5.62) compared to the Standard baseline. This advantage is robust to scaling: at the \textbf{GPT-L} size, while the layer-wise LaVE baseline actually degrades performance relative to the standard model (FID 3.77 vs. 3.47), MoVE maintains a clear lead, achieving the best FID of \textbf{3.10}. These results confirm that the global value embedding bank is an effective primitive for visual modeling, allowing the model to efficiently store and retrieve diverse visual patterns without the instability observed in layer-wise memory at scale.

\subsection{Architectural Versatility: Extension to MLA}

Finally, we demonstrate that MoVE is robust not only across modalities but also across distinct attention mechanisms. We apply MoVE to \textbf{Multi-Head Latent Attention (MLA)} \citep{deepseekv2}, a state-of-the-art architecture that optimizes inference through Key-Value compression.

\textbf{Preliminaries.}
MLA mitigates the memory bottleneck of standard attention by compressing Keys and Values into a low-rank latent vector $\mathbf{c}_{KV}$. Formally, the input $X$ is first projected into the latent space via a down-projection matrix $W_{DKV} \in \mathbb{R}^{d \times d_c}$ such that $\mathbf{c}_{KV} = X W_{DKV}$.
In principle, the full-rank Values ($V$) could be reconstructed via an up-projection $W_{UV} \in \mathbb{R}^{d_c \times d_h H}$. However, to maximize efficiency, MLA absorbs this up-projection into the final output projection matrix $W_O$. The contribution of the values to the final output $O$ is effectively $O \propto (\mathbf{c}_{KV} W_{UV}) W_O = \mathbf{c}_{KV} (W_{UV} W_O)$.
This formulation ensures that the high-dimensional tensor $V$ is never materialized, and only the compact $\mathbf{c}_{KV}$ is stored in the cache.

% FIGURE INSERTION
\begin{figure}[t]
\begin{center}
\centerline{\includegraphics[width=\columnwidth]{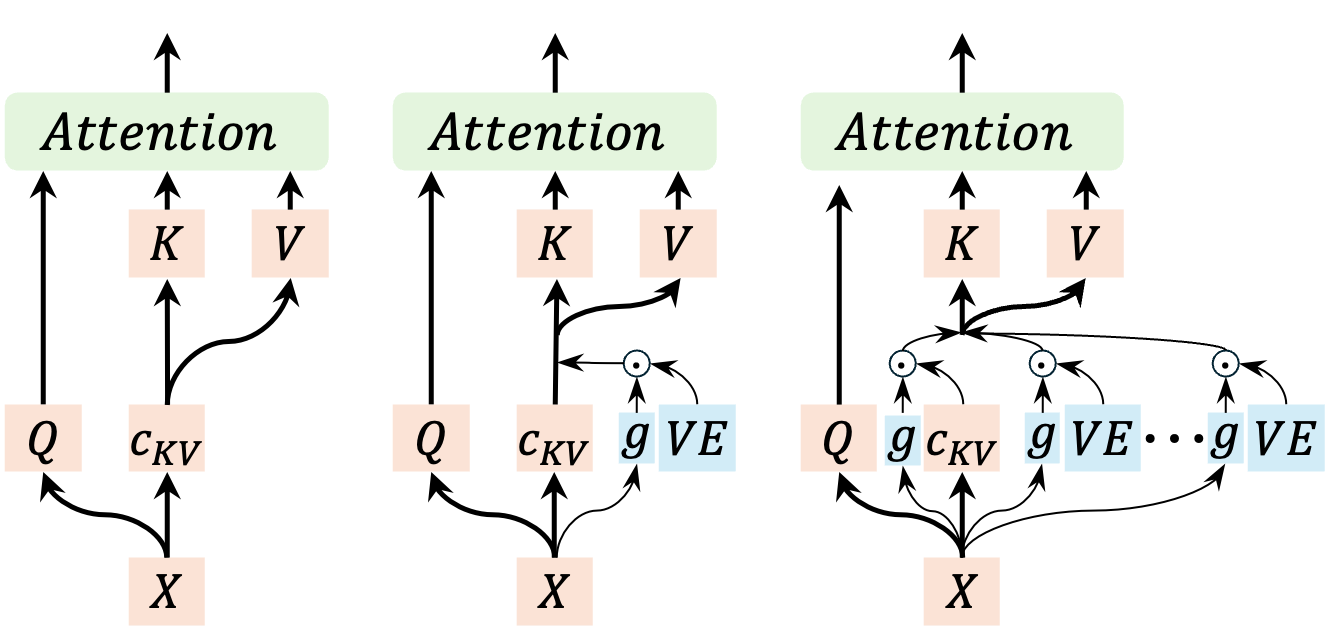}}
\caption{Conceptual Illustration of Memory-Augmented Multi-Head Latent Attention. \textbf{Left:} Standard MLA compresses Key-Value heads into a low-rank latent vector $\mathbf{c}_{KV}$ to minimize cache footprint. \textbf{Center:} MLA + LaVE, which augments the latent space with layer-specific parameters. \textbf{Right:} MLA + MoVE (Ours), which injects a global shared memory into the latent space.}
\label{fig:mla_move}
\end{center}
\end{figure}

\textbf{Memory Injection in MLA.}
We visualize the integration of our method into MLA in Figure \ref{fig:mla_move}. We adapt MoVE to inject memory directly into this efficient latent space. Crucially, to maintain the expressivity of head-specific gating found in MHA, we partition the latent vector $\mathbf{c}_{KV}$ into $H_{kv}$ chunks and apply MoVE head-wise:
\begin{equation}
    \mathbf{c}_{KV, \mathcal{S}}^{(h)} = g_{t, 0}^{(h)} \odot \mathbf{c}_{KV}^{(h)} + \sum_{i=1}^{M} g_{t, i}^{(h)} \odot \mathcal{M}_{t, i}^{(h)}
\end{equation}
where $\mathbf{c}_{KV}^{(h)}$ represents the $h$-th slice of the compressed latent vector and $\mathcal{M}_{t, i}^{(h)}$ corresponds to the retrieved memory for that specific head chunk. The scaled latent vector $\mathbf{c}_{KV, \mathcal{S}}$ (formed by concatenating all $\mathbf{c}_{KV, \mathcal{S}}^{(h)}$) is then used in place of the original $\mathbf{c}_{KV}$ for the final projection. Consistent with the standard MLA formulation, this augmented latent vector $\mathbf{c}_{KV, \mathcal{S}}$ is the specific tensor stored in the KV cache, preserving the architecture's memory compression benefits while embedding the retrieved global knowledge.

\begin{table}[t]
\caption{MoVE applied to Multi-Head Latent Attention (MLA). We compare against a LaVE baseline and perform an extensive sweep on D12 and D20 (up to $\times 32$). Note the \emph{parameter efficiency}: due to the $1/32$ compression ratio, the $\times 32$ setting adds roughly the same number of parameters as a standard $\times 1$ bank.}
\label{tab:mla_scaling}
\begin{center}
\begin{small}
\begin{sc}
\resizebox{\columnwidth}{!}{%
\begin{tabular}{lllrcc}
\toprule
Depth & Method & $M$ & Params & Val BPB $\downarrow$ & Gain $\uparrow$ \\
\midrule
\multirow{9}{*}{D12} & MLA (Base) & - & 172M & 0.8826 & - \\[2pt] \cdashline{2-6} \addlinespace[2pt]
      & \multirow{2}{*}{+ LaVE} & $\times 1$ & +9.5M & 0.8803 & 0.0023 \\
      &                         & $\times 2$ & +18.9M & 0.8806 & 0.0020 \\[2pt] \cdashline{2-6} \addlinespace[2pt]
      & \multirow{6}{*}{\textbf{+ MoVE}} & $\times 1$ & +9.8M & 0.8822 & 0.0004 \\
      &                                  & $\times 2$ & +19.6M & 0.8805 & 0.0021 \\
      &                                  & $\times 4$ & +39.1M & 0.8782 & 0.0044 \\
      &                                  & $\times 8$ & +78.2M & 0.8762 & 0.0064 \\
      &                                  & $\times 16$ & +156.4M & 0.8728 & 0.0098 \\
      &                                  & $\times 32$ & +312.7M & \textbf{0.8690} & \textbf{0.0136} \\
\midrule
\multirow{9}{*}{D20} & MLA (Base) & - & 499M & 0.7868 & - \\[2pt] \cdashline{2-6} \addlinespace[2pt]
      & \multirow{2}{*}{+ LaVE} & $\times 1$ & +26.3M & 0.7857 & 0.0011 \\
      &                         & $\times 2$ & +52.7M & 0.7861 & 0.0007 \\[2pt] \cdashline{2-6} \addlinespace[2pt]
      & \multirow{6}{*}{\textbf{+ MoVE}} & $\times 1$ & +29M & 0.7861 & 0.0007 \\
      &                                  & $\times 2$ & +58M & 0.7853 & 0.0015 \\
      &                                  & $\times 4$ & +115M & 0.7851 & 0.0017 \\
      &                                  & $\times 8$ & +230M & 0.7826 & 0.0042 \\
      &                                  & $\times 16$ & +461M & 0.7808 & 0.0060 \\
      &                                  & $\times 32$ & +921M & \textbf{0.7785} & \textbf{0.0083} \\
\bottomrule
\end{tabular}
}
\end{sc}
\end{small}
\end{center}
\end{table}

\textbf{Motivation.} It is important to note why we operate on $\mathbf{c}_{KV}$ rather than the conceptual $V$. If we were to apply MoVE in the standard manner—injecting memory into the up-projected Value heads—we would be forced to materialize the massive full-rank tensor $V$, effectively \textbf{negating the architectural efficiency of MLA}. The core efficiency of MLA relies on absorbing the up-projection $W_{UV}$ into the output projection $W_O$ (i.e., computing $\mathbf{c}_{KV} (W_{UV} W_O)$). By restricting the memory injection to the compressed latent space, we preserve this absorption, allowing us to scale parametric memory without ever paying the computational or memory price of full-rank tensor materialization.

\textbf{Results.}
Table \ref{tab:mla_scaling} details the performance comparisons. The results reveal a sharp distinction between local and global memory scaling in the compressed latent space.
The \textbf{LaVE} baseline struggles in this efficient architecture; on both D12 and D20, doubling the memory capacity ($\times 1 \to \times 2$) results in slight performance degradation (e.g., $0.8803 \to 0.8806$ on D12). This suggests that strictly coupling memory to layers within a compressed bottleneck saturates quickly.
In contrast, \textbf{MoVE} demonstrates sustained scalability. On \textbf{D12}, it improves monotonically up to the $\times 32$ setting. Notably, due to the aggressive $1/32$ compression ratio of MLA, the MoVE-$\times 32$ configuration adds roughly the same parameter count ($+312.7$M) as a standard uncompressed $\times 1$ bank, yet achieves a substantial gain of \textbf{0.0136} BPB.
On \textbf{D20}, the trend persists: while LaVE saturates, MoVE continues to scale, reaching a BPB of \textbf{0.7785} at $\times 32$. These findings indicate that MoVE and MLA are highly synergistic: MLA provides the inference efficiency (via KV compression), while MoVE expands the model's capacity (via Parametric Memory) without breaking the compute budget.

\subsection{Ablation Study}

MoVE introduces two key structural differences compared to the LaVE baseline: 1) a global, shared memory bank ($\mathcal{E}$), and 2) a learnable gate on the standard value projection ($g_0$). To understand the contribution of each component, we perform targeted ablations on the \textbf{D12} and \textbf{D20} architectures in both the $\times 1$ and $\times 2$ settings:

\textbf{1. MoVE - Gated Standard Path.} We restrict MoVE by fixing the standard path gate to identity ($g_0 = 1$), forcing it to operate as a residual addition: $V_{\mathcal{S}}^{(h)} = V^{(h)} + \sum g_i \odot \mathcal{M}_i$.

\textbf{2. LaVE + Gated Standard Path.} We enhance LaVE by applying a sigmoid gate to the standard projection: $V_{\mathcal{S}}^{(l)} = g_0 \odot V^{(l)} + \dots$. This grants LaVE the same capability to suppress or amplify the input context as MoVE.

\begin{table}[t]
\caption{Ablation study on D12 and D20. We disentangle the benefits of gating the standard path ($g_0$) from the Global Shared Memory architecture. \emph{Val BPB} is reported (\textbf{lower is better}).}
\label{tab:ablation}
\begin{center}
\begin{small}
\begin{sc}
\resizebox{\columnwidth}{!}{%
\begin{tabular}{lll|cc|cc}
\toprule
\multirow{2}{*}{Arch.} & \multirow{2}{*}{Memory} & \multirow{2}{*}{Std. Path} & \multicolumn{4}{c}{Val BPB $\downarrow$} \\
 & & & D12 & D12-$\times 2$ & D20 & D20-$\times 2$ \\
\midrule
\multirow{2}{*}{LaVE} & Local  & Ungated     & 0.8223 & 0.8182 & 0.7526 & 0.7479 \\
                      & Local  & Gated ($g_0$) & 0.8229 & 0.8128 & 0.7514 & 0.7462 \\
\midrule
\multirow{2}{*}{MoVE} & \textbf{Global} & Ungated     & 0.8229 & 0.8176 & 0.7484 & 0.7465 \\
                      & \textbf{Global} & Gated ($g_0$) & \textbf{0.8189} & \textbf{0.8124} & \textbf{0.7478} & \textbf{0.7440} \\
\bottomrule
\end{tabular}
}
\end{sc}
\end{small}
\end{center}
\end{table}

\looseness=-1
The results in Table \ref{tab:ablation} demonstrate that both components contribute positively to model performance. The Global Shared Memory architecture acts as the primary driver of efficiency; MoVE consistently outperforms the enhanced LaVE baseline (e.g., 0.8124 vs 0.8128 on D12-$\times 2$), confirming that decoupling memory from depth provides a distinct advantage beyond simple parameter count. Simultaneously, gating the standard path ($g_0$) proves particularly effective at higher memory densities, offering a substantial boost over the ungated configuration (0.8182 $\to$ 0.8128 on LaVE D12-$\times 2$). We conclude that while the global bank provides the scalable capacity, the gating mechanism is essential for the model to effectively arbitrate between this retrieved knowledge and its local context.
\section{Limitations and Future Work}

\textbf{The Cost of Memory: Bandwidth vs. FLOPs.}
While we have established MoVE as a compute-efficient axis for scaling, it introduces a distinct cost in the form of memory bandwidth. The primary trade-off is \textbf{Compute vs. Memory Access}. Although retrieving and mixing embeddings requires negligible arithmetic overhead, it necessitates fetching parameters from High Bandwidth Memory (HBM). While MoVE effectively allows system designers to ``buy'' model quality using available memory bandwidth rather than compute cycles, determining the optimal trade-off between these resources is highly hardware-dependent. For instance, given that MoVE-D12 significantly outperforms the Standard-D12 baseline, it is intriguing to ask if a memory-augmented shallow model (e.g., D12+MoVE) could match or exceed the performance of a slightly deeper dense model (e.g., D13 or D14) while maintaining a significantly lower FLOP count. Establishing this precise ``exchange rate'' between memory slots and transformer layers would be valuable for identifying optimal architectures. \emph{However, conducting the fine-grained sweep of varying depths required to map this Pareto frontier is resource-intensive and lies outside the scope of this initial study.} We leave the search for these specific architectural ``sweet spots'' to future work.

\textbf{MoVE vs. MoE.}
It is instructive to contrast MoVE with the Mixture-of-Experts paradigm. While both decouple capacity from dense compute, they operate on distinct levels: Standard MoE sparsifies computation at the \textbf{block level} via hard routing, whereas MoVE densifies memory access at the \textbf{parameter level} via soft gating. These approaches offer complementary scaling benefits—MoE efficiently scales reasoning depth, while MoVE scales semantic breadth. We hypothesize that combining MoE for reasoning blocks with MoVE for attention memory could maximize both efficiency and capacity, though we leave the empirical validation of this hybrid architecture to future work.

\looseness=-1
Beyond finding the optimal balance between MoVE and other architectural components, the internal design of MoVE itself is likely far from optimal. We acknowledge that the \textit{parameter efficiency} of our current implementation is noticeably lower than that of the standard model; while MoVE effectively decouples capacity from compute, the performance gain per additional parameter is significantly less than in standard dense scaling. Consequently, improving the mechanism's efficiency in this direction remains a compelling challenge. For instance, could value storage be made more compact by allowing semantically related tokens to share embedding banks? Furthermore, could the computational overhead be minimized by employing a head-wise projection $W_G$ to further reduce routing FLOPs? Our work opens this new axis for scaling parametric memory, and we leave the exploration of these optimizations as fertile ground for future research.
\section{Conclusion}

In this work, we introduced \textbf{MoVE (Mixture of Value Embeddings)} to resolve the structural tension between parametric capacity and computational cost in autoregressive models. By augmenting the attention mechanism with a globally shared, differentiable memory bank, MoVE effectively decouples the scaling of stored knowledge from the depth of reasoning. Unlike traditional layer-wise approaches that saturate quickly, MoVE enables sustained performance improvements even in ``super-dense'' regimes where memory capacity far exceeds layer-wise injection approaches.

Our empirical results span Text Generation, Image Generation, and efficient architectures like MLA, demonstrating consistent gains across diverse modalities. We believe these findings position MoVE as a promising architectural primitive for Generative AI, offering a scalable path toward ``memory-dense'' systems that combine vast encyclopedic capacity with computational efficiency, effectively establishing a new scaling dimension orthogonal to network depth.
\section*{Acknowledgements}

We thank Andrej Karpathy for his continuous efforts in creating high-quality codebases that significantly facilitate research on GPT models. We also thank Peize Sun, Yi Jiang, Shoufa Chen, Shilong Zhang, Bingyue Peng, Ping Luo, and Zehuan Yuan for open-sourcing LlamaGen. Last but not least, we thank the teams at Ant Research Interactive Intelligence Lab and Robbyant for their insightful discussions and infrastructure support.
\section*{Impact Statement}

This paper presents work whose goal is to advance the field of Machine Learning. There are many potential societal consequences of our work, none which we feel must be specifically highlighted here.


\bibliography{references}
\bibliographystyle{icml2026}

%%%%%%%%%%%%%%%%%%%%%%%%%%%%%%%%%%%%%%%%%%%%%%%%%%%%%%%%%%%%%%%%%%%%%%%%%%%%%%%
%%%%%%%%%%%%%%%%%%%%%%%%%%%%%%%%%%%%%%%%%%%%%%%%%%%%%%%%%%%%%%%%%%%%%%%%%%%%%%%
% APPENDIX
%%%%%%%%%%%%%%%%%%%%%%%%%%%%%%%%%%%%%%%%%%%%%%%%%%%%%%%%%%%%%%%%%%%%%%%%%%%%%%%
%%%%%%%%%%%%%%%%%%%%%%%%%%%%%%%%%%%%%%%%%%%%%%%%%%%%%%%%%%%%%%%%%%%%%%%%%%%%%%%
\newpage
\appendix
\onecolumn

\section{Computational Overhead Derivation}
\label{sec:appendix_flops}

In this section, we provide the detailed breakdown of the floating-point operations (FLOPs) required to process a single token within a sequence of length $T$.

\textbf{Baseline Transformer Cost ($C_{std}$).}
A standard Transformer layer comprises three primary computational components:
\begin{enumerate}
    \item \textbf{Dense Projections:} The Query, Key, Value, and Output projections ($W_Q, W_K, W_V, W_O$) operate on the model dimension $d$. Collectively, they consume $4 \times 2d^2 = 8d^2$ FLOPs per token.
    \item \textbf{Feed-Forward Network (FFN):} With a standard expansion factor of 4, the FFN up- and down-projections consume $2 \times 2d(4d) = 16d^2$ FLOPs per token.
    \item \textbf{SDPA Core:} The Scaled Dot-Product Attention mechanism scales quadratically with the sequence length. For a full sequence, the operations $QK^T$ and $Attention(Q,K)V$ require $\approx 4T^2d$ FLOPs, translating to an amortized cost of $4Td$ FLOPs per token.
\end{enumerate}
Summing these yields a baseline budget of:
\begin{equation}
    C_{std} \approx 24d^2 + 4Td
\end{equation}

\textbf{MoVE Overhead ($C_{move}$).}
The computational overhead from MoVE is restricted to the routing projection $W_G \in \mathbb{R}^{d \times H(M+1)}$. This dense projection consumes:
\begin{equation}
    C_{move} \approx 2d \cdot H(M+1)
\end{equation}

\textbf{Relative Ratio.}
Dividing the overhead by the baseline yield the ratio presented in Equation 8:
\begin{equation}
    \text{Ratio} = \frac{2 d H (M+1)}{24 d^2 + 4Td} = \frac{H(M+1)}{12d + 2T}
\end{equation}
Substituting the values from our \textbf{D32} experiment ($d=2048, H=16, M=32, T=2048$):
\begin{equation}
    \text{Ratio}_{D32} = \frac{16(33)}{12(2048) + 2(2048)} = \frac{528}{24576 + 4096} \approx 1.8\%
\end{equation}
This confirms that MoVE breaks the memory-compute coupling, effectively scaling capacity for free ($<2\%$ cost) relative to the dense backbone.

\section{Visualization of MoVE Routing}
\label{sec:appendix_vis}

To investigate whether the D32-MoVE-$\times1$ model utilizes its global memory slots to encode specific semantic features within these constraints, we perform a controlled qualitative study. We analyze the gating behavior of the model when processing a target word in three distinct contexts:
\begin{itemize}
    \item \textbf{Meaning 1 (Control Pair A1 \& A2):} Two sentences where the target word shares the same meaning.
    \item \textbf{Meaning 2 (Contrast B1):} A sentence where the target word has a distinct meaning.
\end{itemize}

We repeat this analysis across three context lengths (\textbf{Short}, \textbf{Medium}, \textbf{Long}) and organize our examples into three linguistic categories.

\paragraph{Visualization Methodology}
For each target word, we extract the gating activation tensor $G \in \mathbb{R}^{L \times M}$, where $L$ is the number of layers and $M$ is the number of memory slots. The value at position $(l, m)$ represents the average gating weight of memory slot $m$ in layer $l$ associated with the target token. To ensure a rigorous comparison, we visualize these activations in two formats:

\begin{enumerate}
    \item \textbf{Raw Gating Magnitude (Top Rows):} We plot the raw gating vectors for the three sentence variants ($A_1, A_2, B_1$). The color scale is fixed to the range $[0, 2.0]$ to reflect the absolute activation strength of the memory slots (where $1.0$ represents the identity baseline).
    
    \item \textbf{Normalized Differential (Bottom Rows):} To isolate the routing shifts triggered by semantic context, we compute the element-wise absolute difference between the gating vectors. 
    \begin{itemize}
        \item \textbf{Control Diff:} $|A_1 - A_2|$ represents the variation within the same semantic meaning (noise floor).
        \item \textbf{Semantic Diff:} $|A_1 - B_1|$ represents the variation triggered by the change in meaning.
    \end{itemize}
    Crucially, these differences are \textbf{normalized} by the \emph{Global Maximum Difference} observed for that specific word across all three context lengths (Short, Medium, Long). This ensures that the heatmaps are directly comparable across different context depths; a value of $1.0$ (Dark Red) represents the maximum routing shift observed for that word in the entire experiment.
\end{enumerate}

\subsection{Experimental Observations}

\textbf{Disclaimer:} We emphasize that while the routing dynamics presented here validate the functional efficacy of the MoVE mechanism in distributing information across global memory slots, the specific linguistic interpretations are \textbf{preliminary}. The analysis is conducted on the D32-MoVE-$\times1$ architecture, which is significantly smaller in parameter count than commercial foundation models and was trained on a subset of the \emph{FineWeb-Edu} dataset. Consequently, while the distinct routing patterns confirm that the mechanism successfully differentiates contexts, the depth and nuance of the semantic associations observed are specific to this model's scale and training distribution. Caution should be exercised when generalizing these specific linguistic mappings to larger, state-of-the-art systems.

The figures below visualize the routing patterns of D32-MoVE-$\times1$. The bottom rows of each block display the \textbf{normalized absolute difference} between the gating vectors. Our analysis reveals the following behaviors:

\subsubsection{Group 1: Standard Polysemy (Noun vs. Noun)}
\begin{figure*}[!htbp]
\centering
\includegraphics[width=0.9\textwidth, keepaspectratio]{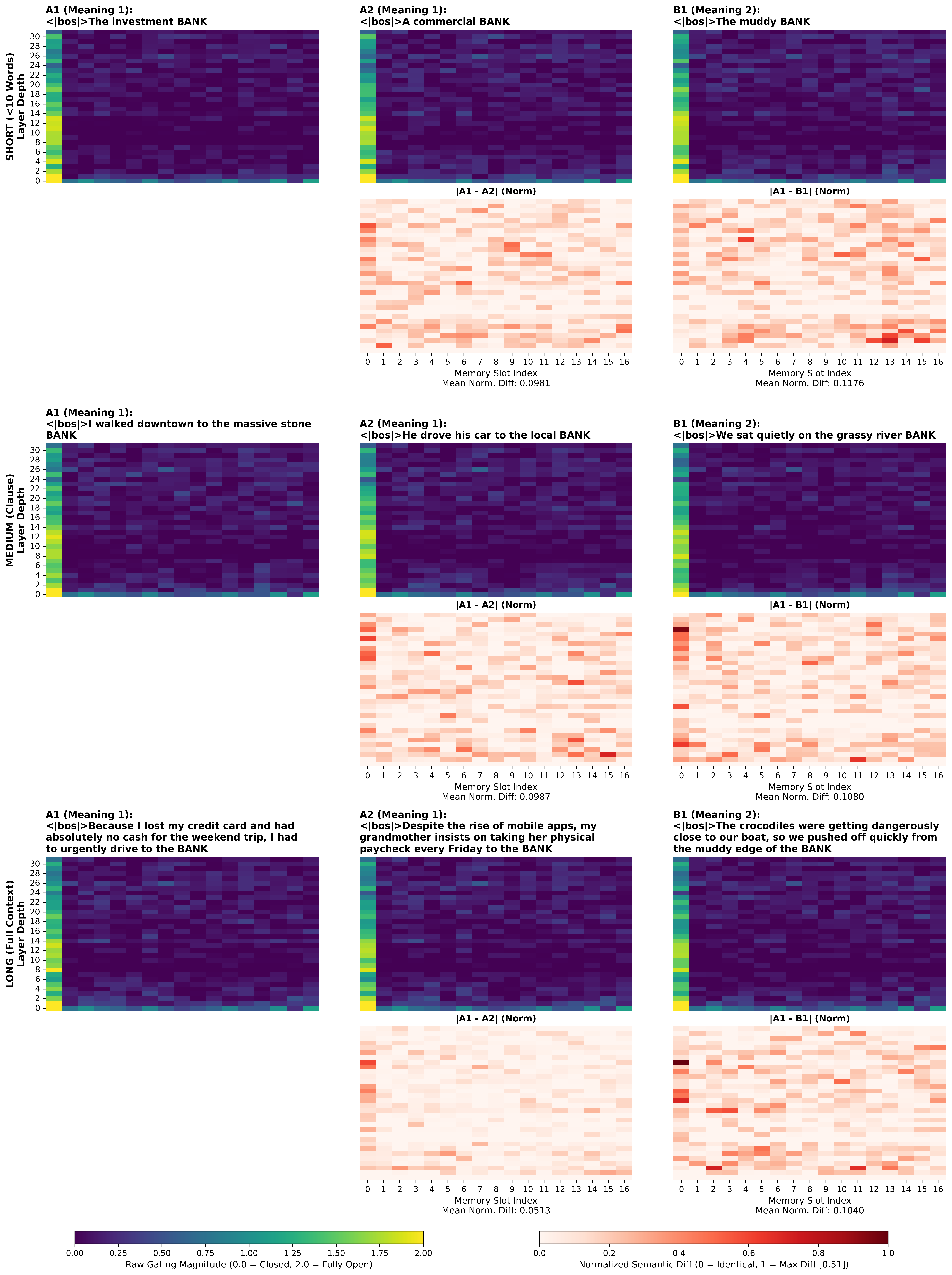}
\caption{Routing visualization for the polysemous word \textbf{``bank''} (Financial Institution vs. River Bank). In the \textbf{Long} context (bottom row), the routing stabilizes: the intra-meaning variation (Control Diff) decreases to 0.0513, while the inter-meaning variation (Semantic Diff) remains high at 0.1040, exhibited by distinct vertical banding.}
\label{fig:vis_bank}
\end{figure*}

\begin{figure*}[!htbp]
\centering
\includegraphics[width=0.9\textwidth, keepaspectratio]{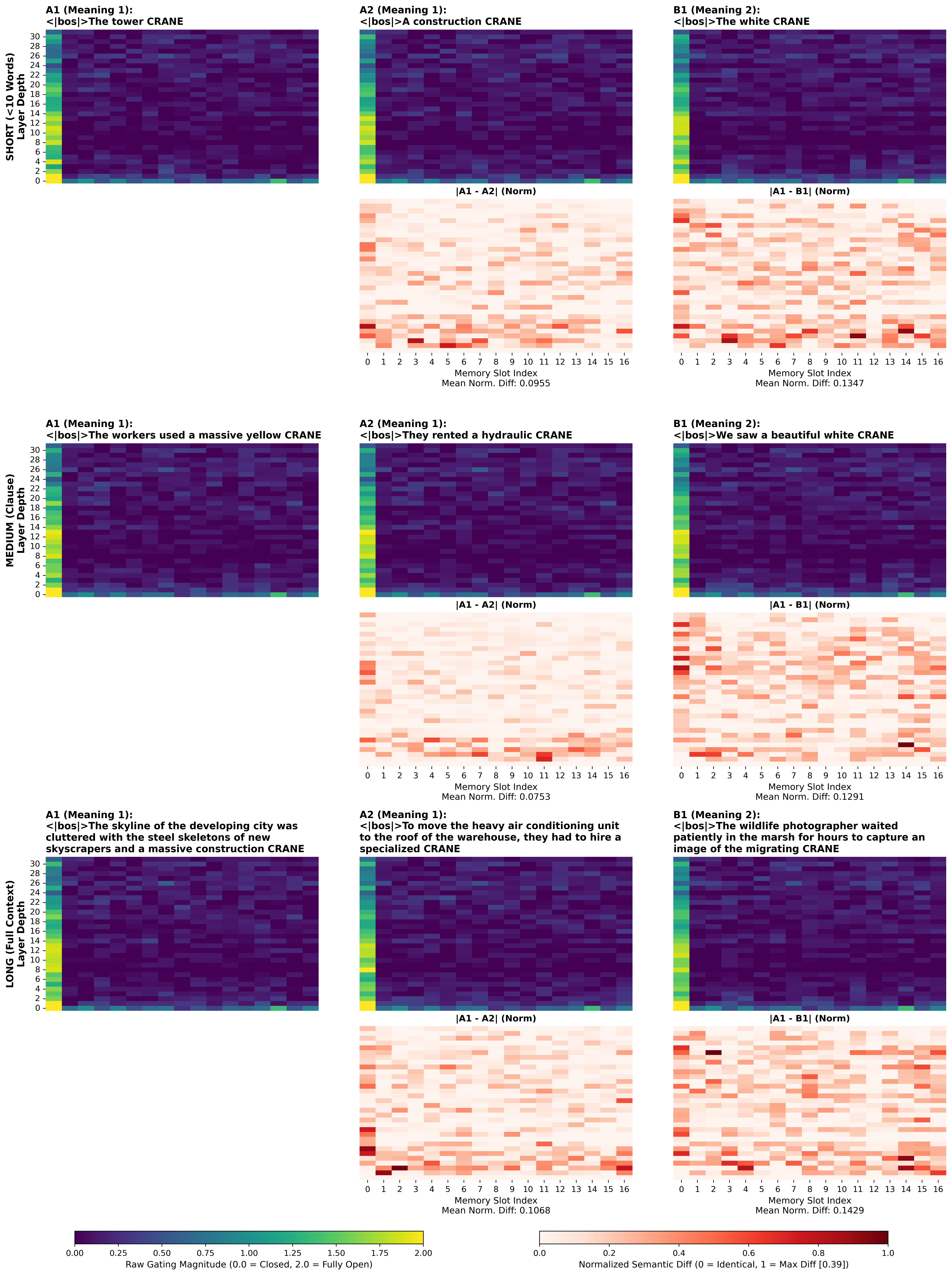}
\caption{Routing visualization for \textbf{``crane''} (Construction Machine vs. Bird). The model maintains a consistent routing separation between the animate and inanimate senses across all context lengths.}
\label{fig:vis_crane}
\end{figure*}

\begin{figure*}[!htbp]
\centering
\includegraphics[width=0.9\textwidth, keepaspectratio]{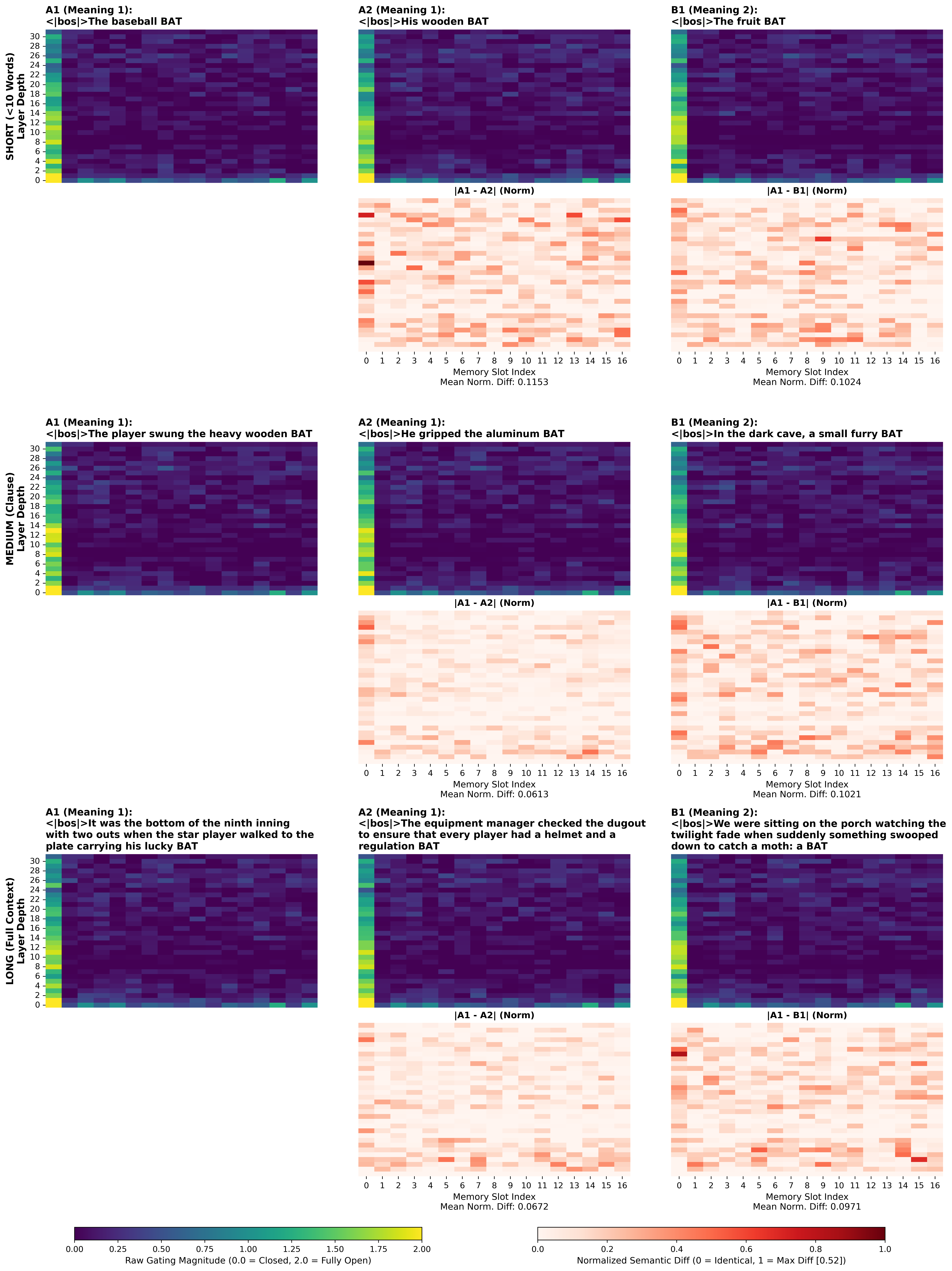}
\caption{Routing visualization for \textbf{``bat''} (Sports Equipment vs. Animal). In the \textbf{Short} context, the routing is noisy: the variation between sentences with the \emph{same} meaning (Control) is actually higher than the variation between \emph{different} meanings (Semantic). However, as context increases (Medium/Long), the signal stabilizes and the Semantic difference correctly dominates.}
\label{fig:vis_bat}
\end{figure*}

\begin{figure*}[!htbp]
\centering
\includegraphics[width=0.9\textwidth, keepaspectratio]{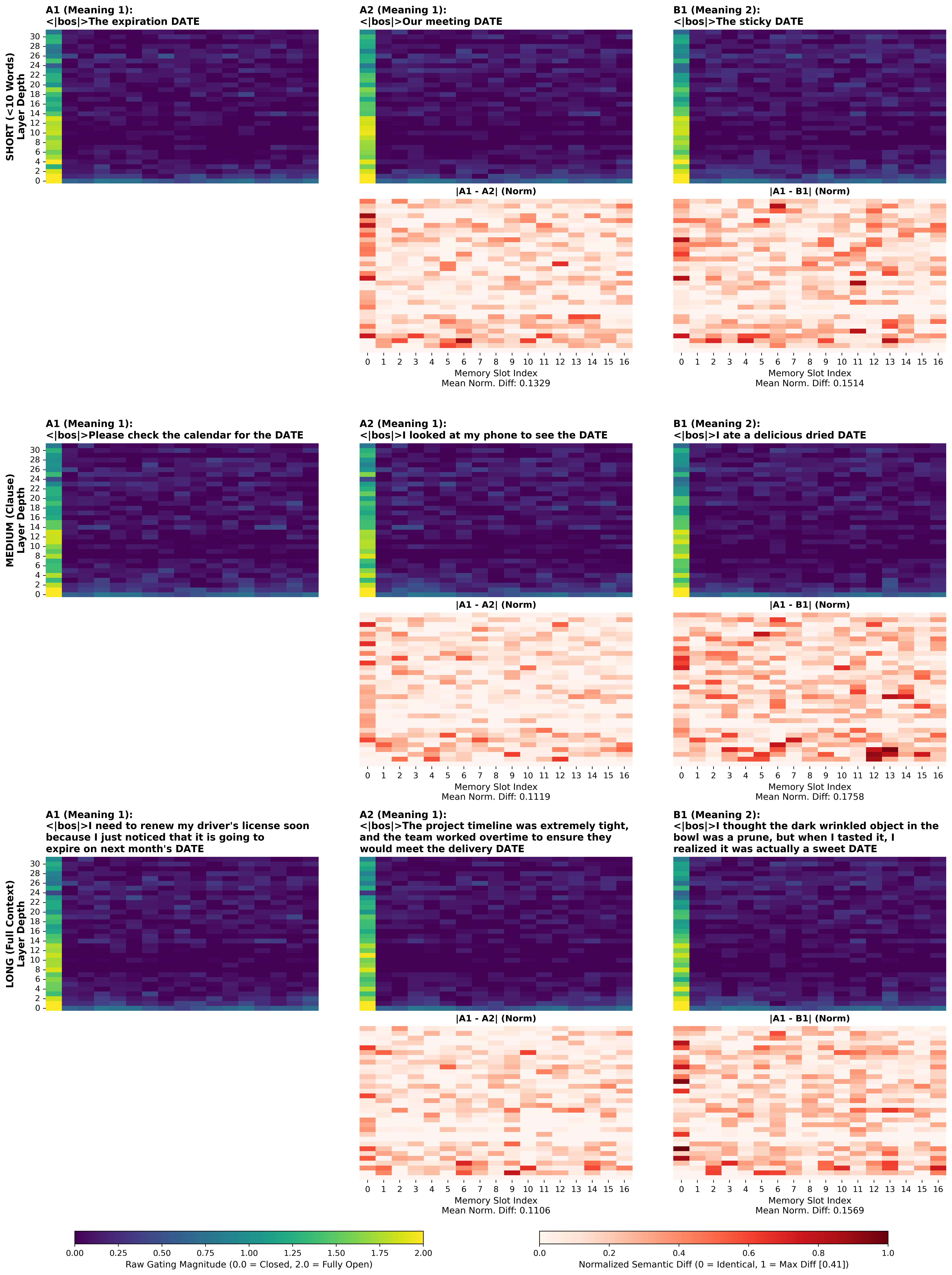}
\caption{Routing visualization for \textbf{``date''} (Calendar Time vs. Fruit). While ambiguous in short contexts, a strong differentiation between the temporal and botanical senses emerges clearly in the Medium context.}
\label{fig:vis_date}
\end{figure*}

In this group (Figures \ref{fig:vis_bank}--\ref{fig:vis_date}), the target word refers to distinct entities.
\begin{itemize}
    \item \textbf{Observation:} In Short contexts ($<10$ words), the routing is often ambiguous, with control and semantic differences showing similar magnitudes (e.g., \emph{Bat} in Figure \ref{fig:vis_bat}).
    \item \textbf{Context Sensitivity:} As context increases to Long (Full Paragraph), a clear signal emerges. The Control difference drops (indicating stability), while the Semantic difference remains high. For \emph{Bank} (Figure \ref{fig:vis_bank}), the Long context reveals distinct ``vertical band'' in the difference heatmap, suggesting that specific memory slots specialize in distinguishing these concepts once sufficient context is available.
\end{itemize}

\subsubsection{Group 2: Syntactic Shift (Noun vs. Verb)}
\begin{figure*}[!htbp]
\centering
\includegraphics[width=0.9\textwidth, keepaspectratio]{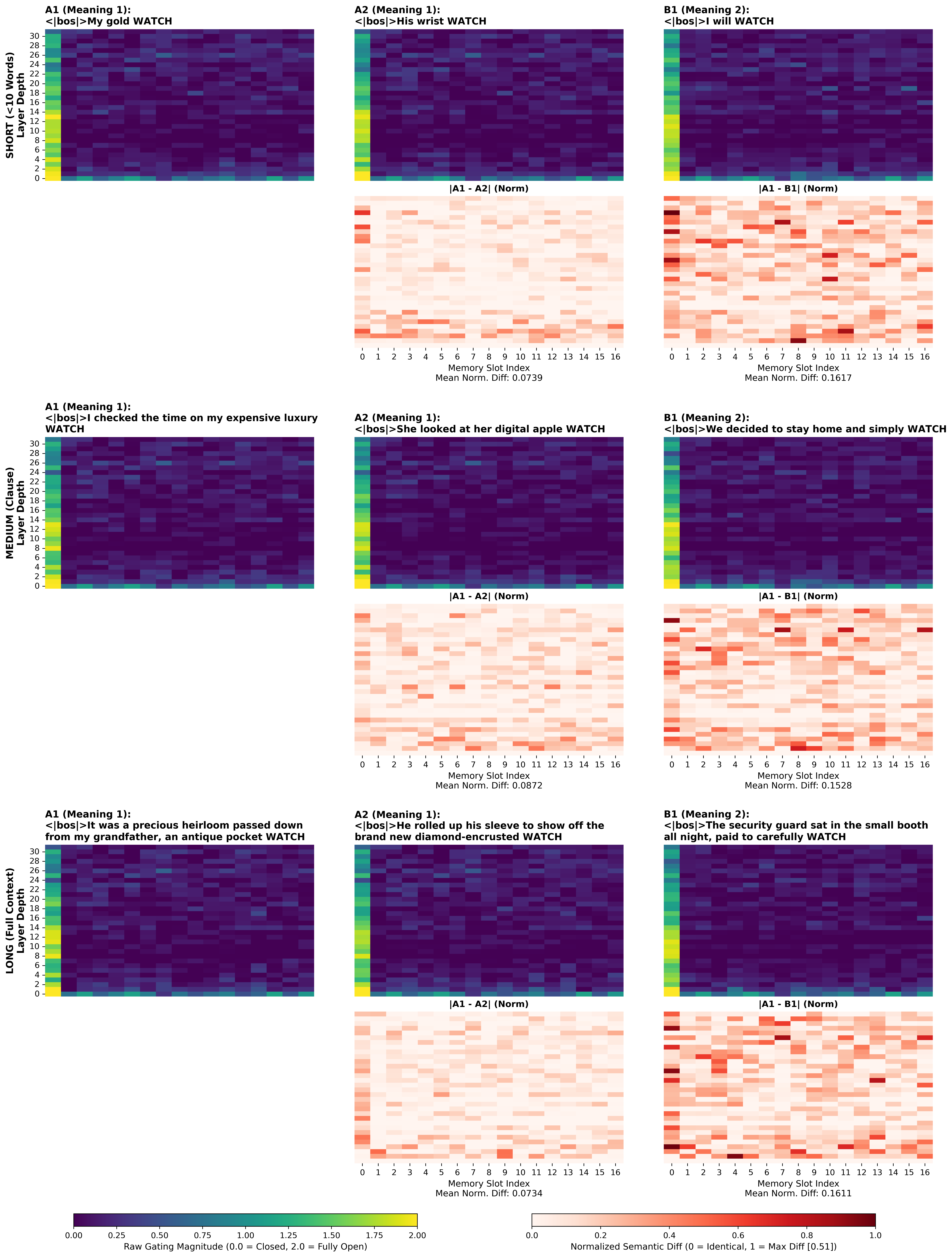}
\caption{Routing visualization for \textbf{``watch''} (Wristwatch/Noun vs. To Look/Verb). This syntactic shift triggers an \textbf{immediate and strong divergence}, even in the Short context. The Semantic difference (0.1617) is nearly double the Control difference (0.0739), suggesting the model relies on local syntax to route verbs differently from nouns.}
\label{fig:vis_watch}
\end{figure*}

\begin{figure*}[!htbp]
\centering
\includegraphics[width=0.9\textwidth, keepaspectratio]{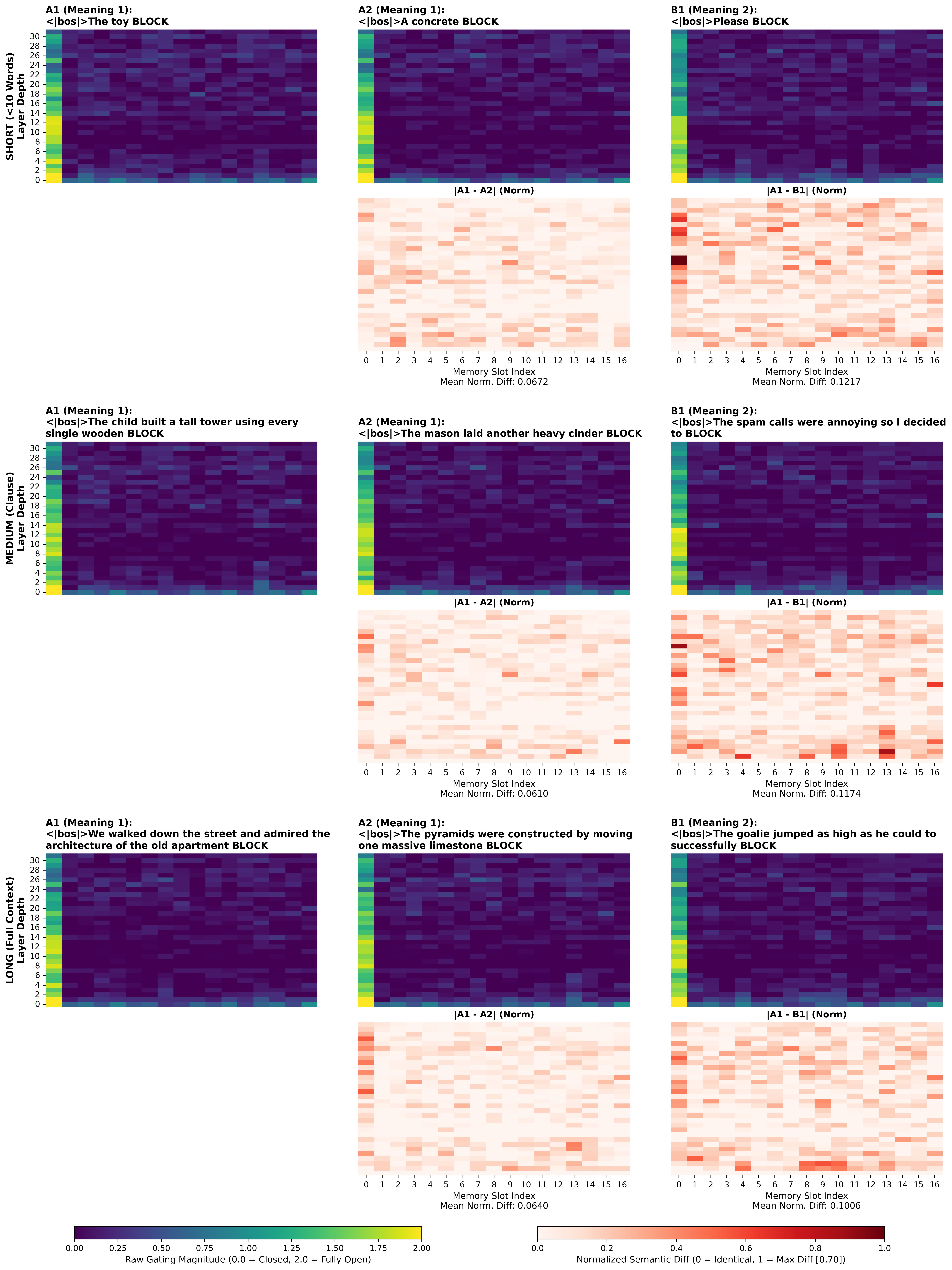}
\caption{Routing visualization for \textbf{``block''} (Toy/Noun vs. Obstruction/Action). Similar to ``watch'', the shift in Part-of-Speech drives a strong routing difference even with minimal context, confirming the mechanism's sensitivity to syntax.}
\label{fig:vis_block}
\end{figure*}

In this group (Figures \ref{fig:vis_watch}--\ref{fig:vis_block}), the word changes grammatical role.
\begin{itemize}
    \item \textbf{Observation:} Unlike polysemy, syntactic shifts trigger \textbf{immediate and strong divergence} in this model. For \emph{Watch} (Figure \ref{fig:vis_watch}), the Short context alone yields a semantic difference (0.1617) nearly double the control difference (0.0739).
    \item \textbf{Implication:} This suggests that the MoVE router is highly sensitive to syntax and Part-of-Speech information, encoding it rapidly in the global memory slots even without long-range context.
\end{itemize}

\subsubsection{Group 3: Metaphorical Shift (Literal vs. Abstract)}
\begin{figure*}[!htbp]
\centering
\includegraphics[width=0.9\textwidth, keepaspectratio]{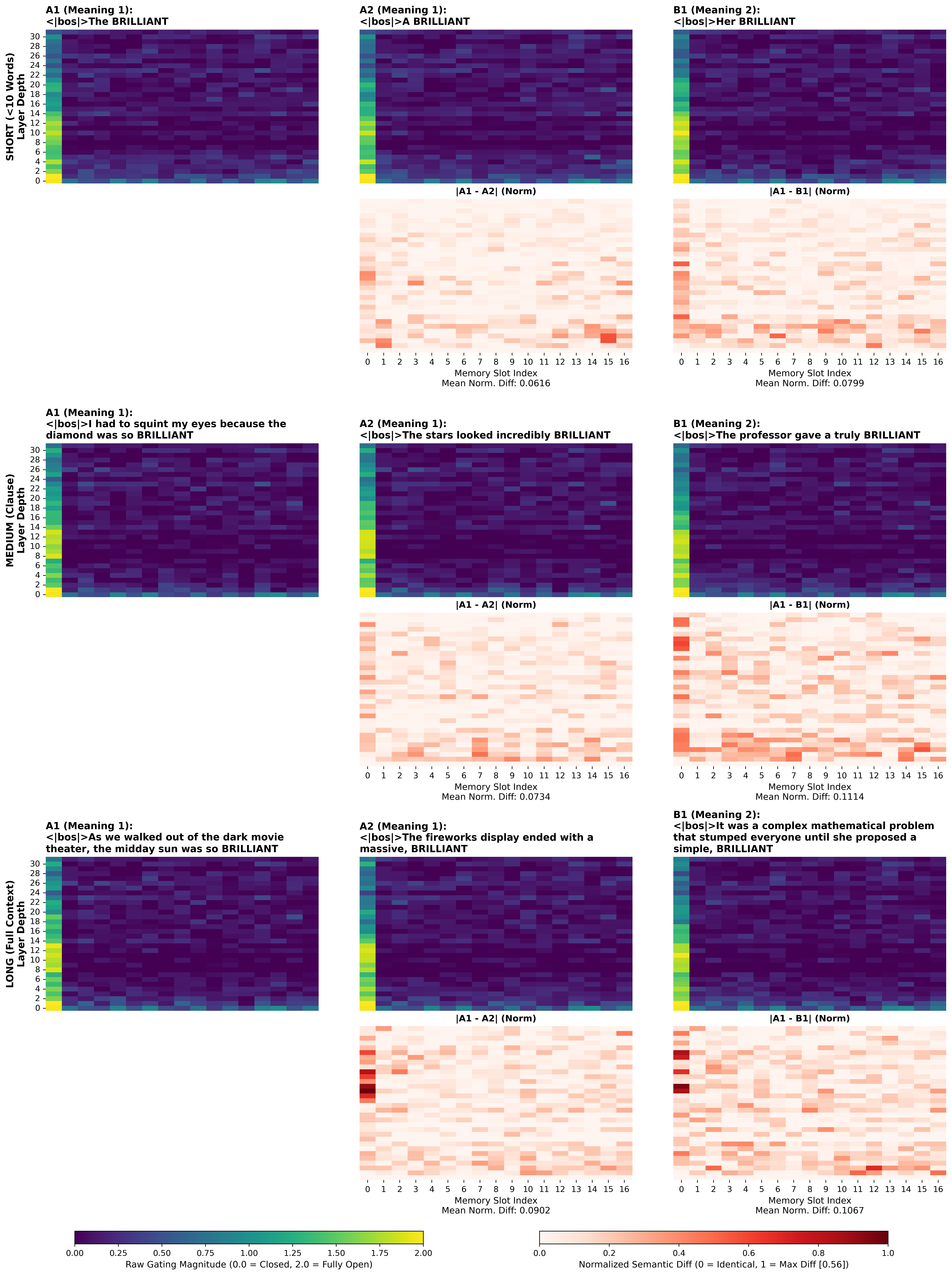}
\caption{Routing visualization for \textbf{``brilliant''} (Literal Light vs. Abstract Intelligence). The metaphorical shift is subtle; distinct routing patterns for the abstract sense become distinct only when sufficient context (Medium and Long) is available.}
\label{fig:vis_brilliant}
\end{figure*}

\begin{figure*}[!htbp]
\centering
\includegraphics[width=0.9\textwidth, keepaspectratio]{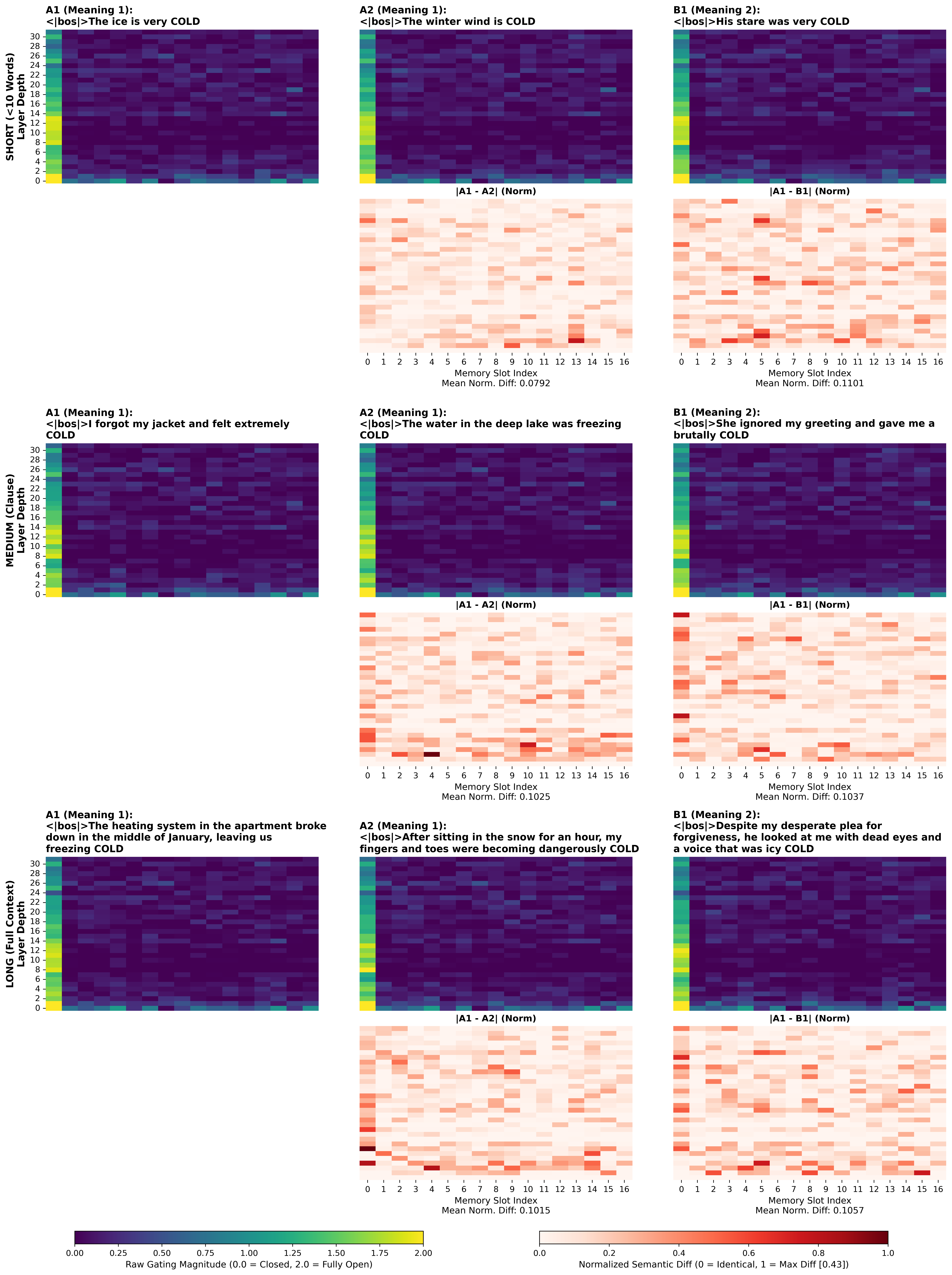}
\caption{Routing visualization for \textbf{``cold''} (Low Temperature vs. Emotionally Distant). The low Semantic Difference (high overlap with Control) suggests the model grounds the abstract metaphorical usage closely to the literal physical sensation, using similar memory slots for both.}
\label{fig:vis_cold}
\end{figure*}

In this group (Figures \ref{fig:vis_brilliant}--\ref{fig:vis_cold}), the word shifts from physical sensation to abstract concept.
\begin{itemize}
    \item \textbf{Observation:} This category exhibits the most subtle gradients. For \emph{Cold} (Figure \ref{fig:vis_cold}), the semantic difference is only marginally higher than the control difference across all context lengths.
    \item \textbf{Implication:} This ``overlap'' supports the linguistic theory of conceptual metaphor, suggesting the model may not treat abstract usages as entirely separate meanings, but rather as extensions grounded in the same semantic slots as their literal roots.
\end{itemize}

\clearpage
\subsection{Category 1: Standard Polysemy}

\clearpage
\subsection{Category 2: Syntactic Shift (Noun vs. Verb)}

\clearpage
\subsection{Category 3: Metaphorical Shift}

%%%%%%%%%%%%%%%%%%%%%%%%%%%%%%%%%%%%%%%%%%%%%%%%%%%%%%%%%%%%%%%%%%%%%%%%%%%%%%%
%%%%%%%%%%%%%%%%%%%%%%%%%%%%%%%%%%%%%%%%%%%%%%%%%%%%%%%%%%%%%%%%%%%%%%%%%%%%%%%

\end{document}